\documentclass[11pt]{article}

\usepackage{microtype}
\usepackage{booktabs}
\usepackage{url}
\usepackage{amsmath}
\usepackage{amsthm}
\usepackage[preprint]{automl}
\usepackage{natbib}
\usepackage{algorithm}
\usepackage{algorithmic}
\usepackage{graphicx}

\title{BOOST: A Data-Driven Framework for the Automated Joint Selection of Kernel and Acquisition Functions in Bayesian Optimization}

\author[1,$\ast$]{\nameemail{Joon-Hyun Park}{redsox45454545@gmail.com}}
\author[1,$\ast$]{\nameemail{Mujin Cheon}{poilk0601@kaist.ac.kr}}
\author[1]{\nameemail{Jeongsu Wi}{wjsoo0907@kaist.ac.kr}}
\author[1,2,3]{\nameemail{Dong-Yeun Koh}{dongyeunkoh@kaist.ac.kr}}

\affil[$\ast$]{Equal contribution.}
\affil[1]{Department of Chemical and Biomolecular Engineering, Korea Advanced Institute of Science and Technology, 291, Daehak-ro, Yuseong-gu, Daejeon, 34141, Republic of Korea}
\affil[2]{Department of AX, Korea Advanced Institute of Science and Technology, 291, Daehak-ro, Yuseong-gu, Daejeon, 34141, Republic of Korea}
\affil[3]{Saudi Aramco-KAIST CO2 Management Center, 291, Daehak-ro, Yuseong-gu, Daejeon, 34141, Republic of Korea}

\hypersetup{%
  pdfauthor={},
  pdftitle={BOOST: A Data-Driven Framework for the Automated Joint Selection of Kernel and Acquisition Functions in Bayesian Optimization},
  pdfsubject={},
  pdfkeywords={Bayesian optimization, Black-box optimization, Active learning, Hyperparameter optimization}
}

\begin{document}

\maketitle

\begin{abstract}
The performance of Bayesian optimization (BO), a highly sample-efficient method
for expensive black-box problems, is critically governed by the selection of its
hyperparameters, including the kernel and acquisition functions.
This presents a significant practical challenge: an inappropriate combination of
these can lead to poor performance and wasted evaluations.
While individual improvements to kernel functions and acquisition functions have
been actively explored, the joint and autonomous selection of the best pair of
these fundamental hyperparameters has been largely overlooked.
This forced practitioners to rely on heuristics or costly manual training.
In this work, we propose a framework, BOOST (Bayesian Optimization
with Optimal Kernel and Acquisition Function Selection Technique), that
automates this selection.
BOOST utilizes a simple offline evaluation stage to predict the performance of
various kernel--acquisition function pairs and identify the most promising pair
before committing to the expensive evaluation process.
BOOST is a data-driven strategy selection procedure that evaluates
kernel--acquisition pairs based on their empirical performance on the
data-in-hand.
At each iteration, previously observed points are partitioned into a reference
set and a query set.
These subsets play roles analogous to training and validation sets in machine
learning: the reference set is used for model construction, while the query set
represents unseen regions to retrospectively evaluate how effectively each
candidate strategy progresses toward the target value.
Experiments on synthetic benchmarks and machine learning hyperparameter
optimization tasks demonstrate that BOOST consistently improves over
fixed-hyperparameter BO and remains competitive with state-of-the-art adaptive
methods, highlighting its robustness across diverse landscapes.
\end{abstract}


\section{Introduction}
Bayesian Optimization (BO) is a sample-efficient framework for optimizing
expensive black-box functions, combining a Gaussian Process (GP) surrogate
with an acquisition function to guide the search
\citep{jones1998efficient,shahriari2015taking,frazier2018tutorial}.
It has been broadly applied to material synthesis
\citep{pruksawan2019prediction,shields2021bayesian}, process optimization
\citep{greenhill2020bayesian,park2024efficient,park2024sustainable}, and
machine learning hyperparameter tuning
\citep{snoek2012practical,klein2017fast,chen2018bayesian,falkner2018bohb,kandasamy2018neural}.

Despite its widespread application, the performance of BO critically depends on
the proper choice of two hyperparameters: the kernel function used in
the GP and the acquisition function guiding the subsequent sample point
\citep{ginsbourger2008discrete,hoffman2011portfolio,snoek2012practical,roman2019experimental,vasconcelos2019no}.
Nevertheless, these hyperparameters are typically fixed throughout the
optimization process, generally chosen arbitrarily or based on standard
defaults, with little attention paid to their selection or adaptation
\citep{genton2001classes,wilson2013gaussian,roman2019experimental}.
This oversight neglects an important insight: the optimal combination of kernel
and acquisition function can vary significantly depending on the structure and
complexity of the objective function.
In practice, such ill-suited configurations slow
convergence, directly inflating the number of evaluations required to
reach optimal targets.
However, the black-box nature of these problems creates a fundamental paradox:
practitioners must select hyperparameters that match the function's
characteristics without knowing those characteristics beforehand.

Existing approaches have attempted to address this gap through adaptive
selection of kernel functions
\citep{ginsbourger2008discrete,malkomes2018automating,roman2019experimental},
acquisition functions \citep{hoffman2011portfolio,vasconcelos2019no}, or
individual development of new kernels
\citep{wilson2016deep,eriksson2021high,jimenez2023scalable,boyne2025bark} and
acquisition functions
\citep{lam2016bayesian,wu2019practical,song2022monte,astudillo2023qeubo,cheon2024earl}.
However, these methods focus on one component while leaving the other fixed,
without considering their joint selection.

To address these challenges, we propose BOOST (Bayesian Optimization with
Optimal Kernel and Acquisition Function Selection Technique)\footnote{Code and data are available at \url{https://figshare.com/s/985f607dccf7ca9505a0}.}, a framework
that identifies the most promising kernel--acquisition pair based entirely on
previously evaluated points (data-in-hand).
Conceptually, the framework adopts a training--validation paradigm:
at each iteration, previously observed points are partitioned into a reference
set and a query set, allowing BOOST to conduct
retrospective evaluations and select the kernel--acquisition pair that most
efficiently rediscovers the optimal points hidden within the query set.
This approach breaks the intrinsic paradox of BO by
leveraging the full structure embedded in the available data, rather than
relying on noisy model uncertainty, converting available data into actionable
insights for robust hyperparameter selection.

Experiments on both synthetic benchmark functions and real-world
hyperparameter optimization tasks show that BOOST consistently outperforms
fixed-hyperparameter baselines, highlighting the importance of careful BO
hyperparameter selection for achieving significant performance gains.

In summary, the key contributions of BOOST are as follows:
\begin{itemize}
    \item Joint selection of kernel and acquisition functions, an aspect largely overlooked in prior work.
    \item Data-driven evaluation using only the data-in-hand, requiring no additional function evaluations and no reliance on model uncertainty estimates.
    \item Iteration-wise adaptation that refines the configuration as new data accumulate.
    \item Empirical validation on synthetic and real-world HPO tasks, showing consistent gains over fixed and adaptive baselines.
\end{itemize}


\section{Preliminaries}
We consider the problem of finding the global minimum
\begin{equation}
x^* = \arg\min_{x \in \mathcal{X}} f(x),
\end{equation}
where $f$ is an expensive black-box function with no access to gradients or structural information. The only available information is a set of observed input--output pairs, the data-in-hand $D_n = \{x_i, f(x_i)\}_{i=1}^n$.

Bayesian Optimization (BO) addresses this challenge by iterating two steps: fitting a Gaussian Process (GP) \citep{williams2006gaussian} surrogate
\begin{equation}
f(x) \sim \mathcal{GP}(m(x), k(x,x'))
\end{equation}
to the data-in-hand, and maximizing an acquisition function $\alpha(x)$ that leverages the GP's predictive mean $\mu(x)$ and uncertainty $\sigma(x)$ to select the next evaluation point
\begin{equation}
x_{n+1} \leftarrow \arg\max_{x \in \mathcal{X}} \alpha(x).
\end{equation}
This process repeats until the evaluation budget is exhausted.

The performance of BO critically depends on two interacting components. The kernel function $k$ encodes structural assumptions about $f$, such as smoothness and characteristic length scales, directly governing the GP's predictive accuracy. In this work, we consider Mat\'ern 3/2, Mat\'ern 5/2, radial basis function (RBF), and rational quadratic (RQ) kernels, which represent different inductive biases.

The acquisition function determines the search strategy by quantifying the utility of candidate points based on the GP's predictive mean and uncertainty. We consider expected improvement (EI), probability of improvement (PI), lower confidence bound (LCB), and posterior mean (PM), which span different degrees of exploration and exploitation. Definitions of both the kernels and acquisition functions are provided in Appendix~\ref{appendix:sec:kernel_acq_defs}.

These components interact: a kernel that underestimates function complexity produces overconfident predictions, which in turn misleads the acquisition function regardless of its design. Conversely, even a well-calibrated surrogate cannot compensate for an acquisition function that is ill-suited to the current optimization stage. Despite this interdependence, most BO studies focus on improving either the kernel or the acquisition function in isolation, and their joint selection remains largely unexplored.


\section{Related Works}
Various approaches have been proposed to address the challenge of selecting
appropriate kernel and acquisition functions in Bayesian Optimization (BO),
recognizing their strong influence on performance.
GP-Hedge \citep{hoffman2011portfolio} addresses the acquisition selection
problem by treating it as a multi-armed bandit, adaptively adjusting the
probability distribution over acquisition functions based on the surrogate
model's predicted value at each candidate point they propose.
However, this approach heavily relies on uncertain, model-driven improvements
from single points, leading to potential instability, especially in the early
stages of optimization.
No-PASt-BO \citep{vasconcelos2019no} mitigates this by incorporating
historical data with a memory factor, yet still suffers from reliance on
uncertain predictive values and does not address kernel adaptation.

Similarly, adaptive kernel selection approaches have been explored.
\citet{ginsbourger2008discrete} proposed using discrete mixtures of kernels
within Gaussian Process-based optimization (originally referred to as Kriging),
where the surrogate model is constructed from a weighted combination of
multiple kernels.
\citet{malkomes2018automating} proposed an automated framework that explicitly
maintains a posterior over multiple kernel structures and selects evaluation
points by averaging the Expected Improvement under this posterior.
\citet{roman2019experimental}, on the other hand, developed a strategy to
select the best-performing kernel among multiple candidates by evaluating them
in parallel, though relying primarily on local surrogate information.
However, while these approaches effectively mitigate kernel misspecification
risks, they typically focus exclusively on the surrogate model while
neglecting the influence of acquisition function choice.
This oversight limits their overall robustness, as the synergy between the
model and the acquisition strategy is crucial for efficiently navigating
complex optimization landscapes.

In contrast to these approaches that focus on individual components, some
recent work has recognized the need for joint selection.
The fundamental challenge in black-box optimization is that practitioners must
choose appropriate kernel and acquisition functions that match the function's
characteristics without knowing those characteristics beforehand, making
simultaneous optimization of both components particularly difficult.
To address this issue, \citet{xue2016accelerated} proposed a data-based
strategy that utilizes the data-in-hand to jointly select the most appropriate
surrogate model and acquisition function for material design tasks.
While this approach represents a significant step toward breaking the
conventional paradox of selecting hyperparameters without prior knowledge, by
transforming available data into actionable insights, their method selects
hyperparameters only at the beginning and does not update them during
subsequent iterations.
This static approach fails to leverage the additional information that becomes
available as the optimization progresses, limiting its adaptability to the
evolving understanding of the objective function.

Building upon this foundation, BOOST introduces an adaptive data-based strategy
that utilizes all previously evaluated points to retrospectively assess
candidate configurations at each iteration.
Unlike previous adaptive kernel function or adaptive acquisition function
studies, BOOST provides several key advantages: it requires no additional
function evaluations, pretraining, or assumptions about model uncertainty for
joint hyperparameter selection.
Moreover, by updating hyperparameter choices at every iteration using the
continuously expanding data-in-hand, BOOST successfully addresses the
conventional paradox of selecting kernel and acquisition functions in
black-box settings, resulting in superior performance.


\section{Methods}

This section outlines the structure of BOOST and how it leverages observed data
to guide BO hyperparameter selection. We describe the overall procedure,
followed by the details of each component. Without loss of generality,
BOOST assumes minimization problems throughout this section.

\subsection{Overall Process of BOOST}
\begin{figure}[t]
\centering
\includegraphics[width=\textwidth]{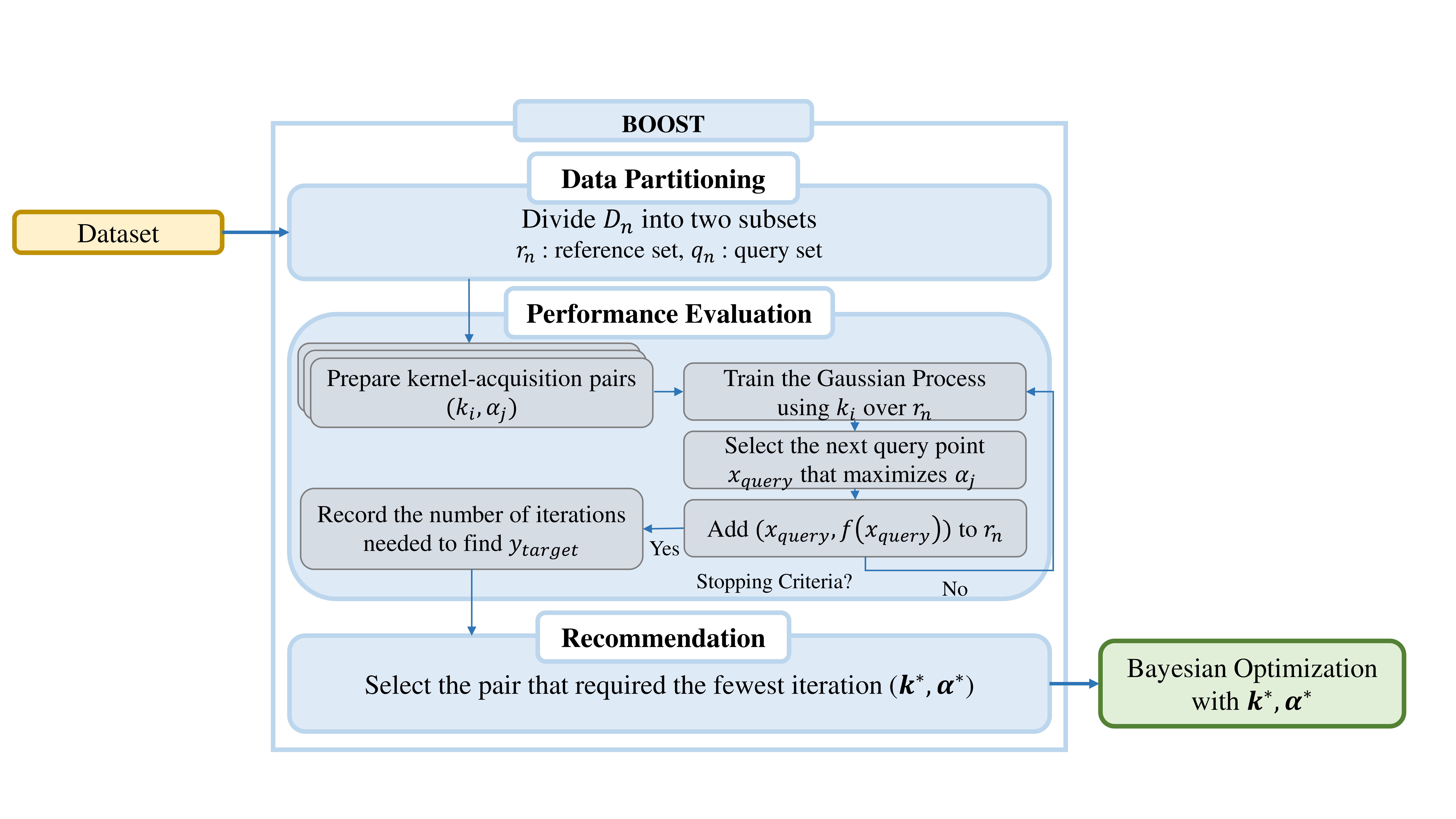}
\caption{Overview of the BOOST architecture. The observed data are first divided into reference and query sets. Candidate kernel--acquisition pairs are then evaluated through an internal BO simulation, and the pair that reaches the target most efficiently is used in the next BO iteration.}
\label{fig1:methods}
\end{figure}

BOOST performs offline evaluation using previously observed points
(data-in-hand) to identify the most promising kernel--acquisition pair from
candidates predefined by practitioners.
BOOST operates in three main steps:
\begin{enumerate}
    \item \textbf{Data Partitioning}: The data-in-hand, $D_n$, is partitioned into two disjoint subsets using K-means clustering. The data points closest to each cluster center serve as the initial subset for constructing a GP (reference set, $r_n$), and the remainder serves as unexplored regions where new query points are generated (query set, $q_n$).
    
    \item \textbf{Performance Evaluation}: For each candidate in kernel--acquisition pairs, BOOST conducts internal BO runs starting with $r_n$. The optimization performance is evaluated based on how many iterations have been required to find the target value, $y_{target}$, within $q_n$.
    
    \item \textbf{Recommendation}: The kernel--acquisition pair that achieves the fastest convergence (i.e., reaches the target value in the fewest steps) is selected for the actual BO process.
\end{enumerate}
The overview of BOOST is shown in Figure~\ref{fig1:methods} and
Algorithm~\ref{alg:boost} in Appendix~\ref{appendix:sec:algorithm}.

\subsection{Data Partitioning}
The data-in-hand, $D_n$, is partitioned into two disjoint subsets: $r_n$ and
$q_n$.
In BOOST, $r_n$, which is the reference subset for the internal BO process,
corresponds to the initial dataset used in the actual BO loop and is utilized
to construct the GP at the beginning of the performance evaluation step.

To ensure that $r_n$ consists of informative and diverse samples while
excluding points already better than the target value (or stopping criteria),
$y_{target}$, we apply K-means clustering on
$D_n \setminus \{(x, f(x)) \mid f(x) \leq y_{target}\}$ to select
representative samples that provide diverse coverage of the promising regions.
From each cluster, the data point closest to the cluster center is selected,
providing both representativeness and coverage of promising regions.
In this study, we define the target value $y_{target}$ for stopping as the top
5th percentile of function values in $D_n$, while the size of $r_n$ is
determined as $\lfloor |D_n| / 3 \rfloor$ with the constraint of
$3 \leq |r_n| \leq 20$ to balance the size of $r_n$ and $q_n$, where
$q_n = D_n \setminus r_n$.
The performance impact of different stopping criteria and data partitioning
methods is analyzed in Appendix~\ref{appendix:sec:ablation}.

All function values in $q_n$, $\{f(x) \mid x \in q_n\}$, are treated as
undiscovered data to assess performance, mimicking the unexplored regions
typically encountered in realistic black-box optimization scenarios.

\subsection{Performance Evaluation}

BOOST evaluates combinations from predefined candidate sets:
\begin{itemize}
    \item Kernel functions: \{Matérn 3/2, Matérn 5/2, RBF, RQ\}
    \item Acquisition functions: \{EI, PI, LCB ($\beta = 0.1$), PM\}
\end{itemize}
The mathematical definitions of these functions are provided in Appendix~\ref{appendix:sec:kernel_acq_defs}.

For each kernel--acquisition pair $(k_i, \alpha_j)$, BOOST performs the following steps:
\begin{enumerate}
    \item Train a Gaussian Process using kernel \( k_i \) over the reference subset \( r_n \).
    
    \item Using predicted mean value and uncertainty from GP in step 1, select next query point \( x_{n+1} \) from \( q_n \) that maximizes acquisition function \( \alpha_j \), and move \( (x_{n+1}, f(x_{n+1})) \) from $q_n$ to $r_n$.
    
    \item Repeat internal BO (i.e., step 1 and step 2) until:
    \begin{itemize}
        \item The target value \( y_{target} \) is reached, i.e., \( f(x_{n+1}) \leq y_{target} \), or
        \item A maximum of 20 iterations is reached.
    \end{itemize}
    
    \item Record the number of iterations \( t_{k_i, \alpha_j} \) required to achieve the target.
    
    \item Repeat steps 1--4 for all other kernel--acquisition combinations.
\end{enumerate}

The internal BO horizon is capped at $T_{\max}=20$, as strategies failing to reach 
the top 5th percentile target within this budget offer no efficiency gain over random search. 
Additional analysis is provided in Appendix~\ref{appendix:sec:ablation}.

The data-based evaluation in BOOST is fully parallelizable, as each
kernel--acquisition pair is assessed independently.
In the context of expensive black-box optimization, where function evaluations often span hours 
to days, the computational overhead of BOOST (on the order of seconds) is negligible relative 
to the time cost of the actual experiments \citep{snoek2012practical,cosenza2022multi}.
Detailed runtime analysis is provided in Appendix~\ref{appendix:sec:time}.

\subsection{Recommendation}
After evaluating all candidate combinations, BOOST selects the pair that required the fewest iterations to reach the target value:
\[
(k^*, \alpha^*) = \arg\min_{(k_i, \alpha_j)} t_{k_i,\alpha_j}
\]
In cases where multiple combinations achieve the same minimum number of iterations, particularly when $D_n$ is small, BOOST breaks the tie using a predefined priority order:
\begin{itemize}
    \item Acquisition functions: EI $>$ PI $>$ LCB $>$ PM
    \item Kernel functions: Matérn 3/2 $>$ Matérn 5/2 $>$ RBF $>$ RQ
\end{itemize}
This priority is based on the degree of exploration each acquisition function
promotes.
When data is scarce, exploration is essential to acquire informative samples
that help reveal the structural characteristics of the objective function.
This, in turn, enables the internal BO process to more effectively evaluate and
distinguish between different kernel--acquisition combinations.
Accordingly, we prioritize acquisition functions with stronger exploratory
tendencies, ranking them as EI $>$ PI $>$ LCB $>$ PM.
In particular, LCB is placed below PI because we adopt a low exploration
coefficient $(\beta = 0.1)$, which makes it behave almost greedily
\citep{dogan2022bayesian,tian2024boundary,papenmeier2025exploring}.

Similarly, we prioritize Matérn kernels over RBF due to their greater
flexibility in modeling non-smooth or complex objective functions, as they are
recommended for many realistic problems where strong smoothness assumptions are
unrealistic \citep{stein1999interpolation,williams2006gaussian,snoek2012practical}.
We place RBF over RQ since RBF is the most widely used kernel, while RQ is a
special case of kernels which can be written as a mixture of RBF kernels with
various length scales \citep{williams2006gaussian}.

The tie-breaking rule incorporates domain-inspired heuristics.
However, the core selection mechanism in BOOST is data-driven, based on
internal BO performance evaluation.
The heuristic component is only applied after quantitative evaluation, and only
when multiple options are equally optimal.
This minimal use of heuristics helps guide the decision in underdetermined
cases without compromising the overall principled framework of BOOST.

\section{Experiments}

We conduct comprehensive experiments to evaluate BOOST's performance across
various discrete optimization scenarios.
All benchmark tasks are designed in fully discretized spaces to align with
practical problem settings and to demonstrate BOOST's real-world applicability
\citep{gonzalez2024survey}.
We compare BOOST against fixed kernel--acquisition function combinations, a
state-of-the-art method, and adaptive approaches on both synthetic benchmark
functions and real-world hyperparameter optimization tasks.

All optimization problems in this section are minimization problems. 
Without loss of generality, maximization objectives were converted to minimization problems by multiplying the objective function values by -1.

\subsection{Synthetic Benchmark Functions}
To assess the robustness of BOOST across diverse optimization landscapes,
we evaluate BOOST on three widely used 4-dimensional discrete synthetic
benchmark functions: Ackley, Levy, and Rosenbrock
\citep{simulationlib}.
Detailed discretization parameters are provided in
Appendix~\ref{appendix:sec:benchmark}.
We use Latin Hypercube Sampling (LHS) to select 10 initial points, ensuring
well-distributed coverage of the search space.
All experiments are repeated 30 times with different sets of initial points to ensure statistical reliability.

\subsection{Real-world machine learning hyperparameter optimization tasks}
To validate the practical applicability of BOOST to real-world machine learning problems,
we conduct
experiments on eight hyperparameter optimization search spaces selected from the HPO-B
dataset \citep{arango2021hpo}.
These search spaces cover Decision Trees (6D), SVM (8D), Random Forests (9--10D),
Bagging + Random Forest (15D), and XGBoost (16D).
For each task, we randomly select 10 initial points from the available
dataset, as the search space is predefined and sampling methods such as LHS are
not applicable.
Detailed experimental configurations and data preprocessing steps are provided
in Appendix~\ref{appendix:sec:hpob}.
All experiments are repeated 30 times with different sets of initial points to ensure statistical reliability.

\subsection{Experimental Setup}
To demonstrate that BOOST consistently outperforms any single fixed strategy,
we first compare BOOST against 16 deterministic baseline methods,
representing all possible combinations of the four kernel functions
(Matérn 3/2, Matérn 5/2, RBF, RQ) and four acquisition functions
(EI, PI, LCB, PM) used in our candidate set.

Next, to demonstrate the importance of jointly optimizing both the kernel and
acquisition function, we compare BOOST against adaptive methods that focus on
selecting only one component while fixing the other. Specifically, we evaluate
Best Utility \citep{roman2019experimental}, an adaptive kernel method, and
No-PASt-BO \citep{vasconcelos2019no}, an adaptive acquisition method.
Additionally, to explicitly verify the impact of the joint selection mechanism
within our framework, we conducted an internal ablation study by fixing either
the kernel or the acquisition function of BOOST. The detailed results and
analysis are provided in Appendix~\ref{appendix:sec:fixed_keracq}.
Beyond these component-wise comparisons, we include Random Search as a model-free baseline. We further evaluate HEBO \citep{cowen2022hebo}, a state-of-the-art black-box optimization algorithm, and qLogNoisyExpectedImprovement (qLogNEI, $q=1$), the default setting of BoTorch \citep{balandat2020botorch}.
We also include Random-KA, which randomly selects a kernel--acquisition pair at each iteration from the same candidate pool, to distinguish BOOST's data-driven selection from the effect of random switching among candidate strategies.

Furthermore, to assess the necessity of adaptive selection at every iteration,
updating the strategy as new data accumulate, we introduce a ``Static''
baseline inspired by \citet{xue2016accelerated}. Similar to BOOST, this method
selects the best kernel--acquisition pair based on the initial data but keeps
this configuration fixed throughout the remaining optimization process.
We also include BOOST-FixedHP, which reuses GP hyperparameters fitted on the data-in-hand when constructing internal GPs. Since BOOST refits GP hyperparameters on each reference subset by default, this ablation isolates how much of the selection benefit comes from adapting GPs in the internal evaluation.


\begin{figure}[t]
\centering
\includegraphics[width=\textwidth]{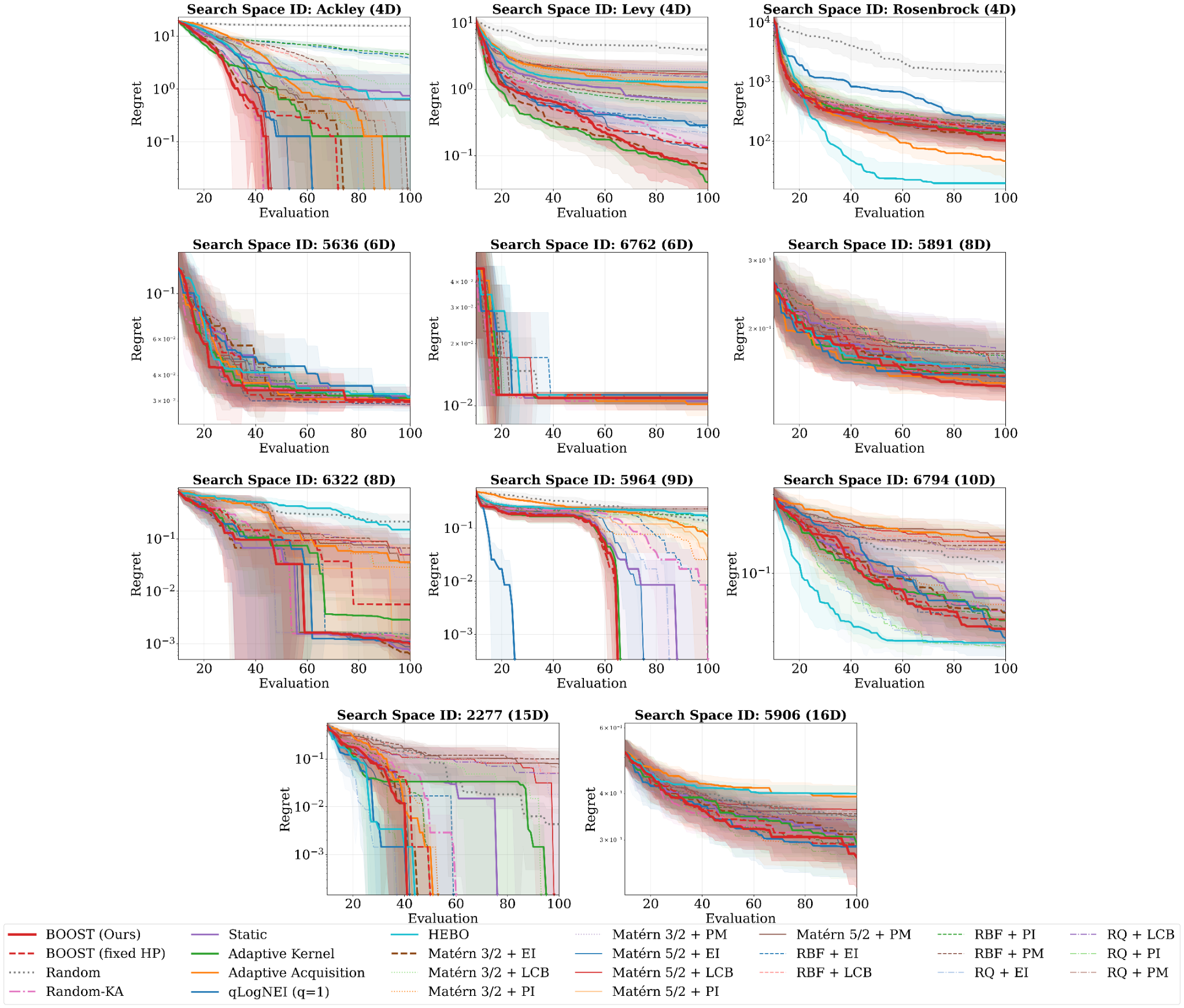}
\caption{Regret curves of BOOST, fixed kernel--acquisition methods, adaptive baselines, and additional external or diagnostic baselines across three synthetic benchmark functions (Ackley, Levy, Rosenbrock; 4D) and eight HPO-B search spaces (2277, 5636, 5891, 5906, 5964, 6322, 6762, 6794; 6--16D). Each experiment uses 10 initial points followed by 90 optimization iterations (100 total evaluations). The x-axis begins at evaluation 10, the point at which all initial observations have been collected and the BO loop starts. Lines show the mean regret over 30 independent runs; shaded regions indicate the 95\% confidence interval.}
\label{fig2:results}
\end{figure}

\subsection{Results}

Figure~\ref{fig2:results} presents the regret curves of all compared methods
across 11 benchmark tasks (3 synthetic functions and 8 HPO-B search spaces).
For a more detailed comparison, the Appendix provides box plots of
the final regret (Figure~\ref{figS1:final_performance}), remaining gap
curves (Figure~\ref{figS2:remaining_gap}), and
worst-case (90th percentile) regret analysis
(Figure~\ref{figS3:worst_regret}). Here, the gap
\citep{huang2006global,wu2019practical} is defined as
$(f^-_{\text{init}} - f^-) / (f^-_{\text{init}} - f^*)$, where
$f^-_{\text{init}}$ is the best value among the initial points, $f^-$ is the
current best observation, and $f^*$ is the global minimum; the remaining gap
($1 - \text{gap}$) quantifies how much of the optimization progress remains
relative to the initial design.

\paragraph{Overall performance}
Across all 11 tasks, BOOST consistently achieves top-tier regret, ranking
among the best methods in every search space (Figure~\ref{fig2:results}).
In contrast, fixed kernel--acquisition pairs tend to perform well on some
search spaces but are strongly influenced by the type of search space, often
failing on others. BOOST avoids such catastrophic failures by adapting its
configuration to the structure revealed by the data-in-hand. For example,
compared with Mat\'ern 3/2 + EI, one of the most widely used combinations, BOOST
shows more stable and consistent performance across tasks. Moreover, even on
search spaces where kernel--acquisition choice has limited impact and most
methods perform similarly (e.g., 5636, 6762), BOOST remains competitive with the
top performers.

Although HEBO and qLogNEI excel on several search spaces, their rankings vary
substantially across tasks, whereas BOOST delivers more consistent regret and
robust worst-case behavior across heterogeneous optimization problems
(Figure~\ref{figS3:worst_regret}). The BOOST-FixedHP comparison suggests
that adaptive hyperparameter refitting improves BOOST's assessment of
kernel--acquisition suitability, supporting more reliable selection.
The comparison with Random-KA demonstrates BOOST's advantage over unstructured random switching among kernel--acquisition choices, indicating that its data-driven selection provides benefits beyond strategy diversity alone.

\paragraph{Comparison with adaptive and static baselines}
Compared with the Static baseline, which selects the best
kernel--acquisition pair from the initial data but fixes it thereafter, BOOST
demonstrates clear gains, confirming the value of iteration-wise adaptation
as new data accumulate (Figure~\ref{fig2:results}). BOOST also outperforms
both Adaptive Kernel \citep{roman2019experimental} and Adaptive Acquisition
\citep{vasconcelos2019no}, which adapt only one component while leaving the
other fixed. Additional ablations in Appendix~\ref{appendix:sec:fixed_keracq}
show that fixing either the kernel or the acquisition function to its
individually best-performing choice still underperforms the jointly adaptive
BOOST. These results validate the core thesis of our work: jointly and
adaptively selecting both the kernel and acquisition function yields superior
performance compared to optimizing either component in isolation.

\section{Conclusion}
In this study, we propose BOOST, a robust framework for selecting appropriate
hyperparameters in Bayesian Optimization (BO).
Motivated by the insight that the data-in-hand can help us to understand the
shape and complexity of the objective function, BOOST performs an internal BO
loop on the available data to identify the kernel and acquisition function
combination that is best suited to the target problem.

Unlike fixed hyperparameter methods that may excel in specific settings but
fail in others, BOOST dynamically adapts its strategy based on observed data
characteristics, eliminating the need for manual tuning while ensuring
consistent performance across diverse optimization scenarios.
While BOOST introduces a few seconds of computational overhead per iteration
(Appendix~\ref{appendix:sec:time}), this cost remains negligible for expensive
black-box evaluations that typically take minutes or longer. In this sense,
BOOST functions as an efficient safeguard against
catastrophic convergence delays caused by ill-suited hyperparameter
configurations, consistently achieving top-tier optimization performance and mitigating
worst-case outcomes across diverse optimization scenarios.

\paragraph{Limitations}
BOOST currently selects from a predefined candidate set of kernel--acquisition
pairs. While we use standard kernels and acquisition functions in this work,
the framework is designed to be extensible: practitioners can incorporate any
kernel or acquisition function suited to their domain, including composite
kernels or custom acquisition strategies.
Our empirical evaluation focuses on discretized search spaces with
low-to-moderate dimensionality (up to 16D); scalability to higher-dimensional
or continuous problems remains to be investigated.
Finally, although the internal retrospective evaluation is lightweight relative
to expensive black-box objectives, it introduces additional computation at each
iteration compared with fixed-strategy BO.


\bibliography{automl2026}

@book{williams2006gaussian,
  author    = "Williams, Christopher K. I. and Rasmussen, Carl Edward",
  title     = "Gaussian Processes for Machine Learning",
  year      = 2006,
  address   = "Cambridge, MA",
  publisher = "MIT Press"
}

@book{stein1999interpolation,
  author    = "Stein, Michael L.",
  title     = "Interpolation of Spatial Data: Some Theory for Kriging",
  year      = 1999,
  address   = "New York",
  publisher = "Springer Science \& Business Media"
}

@article{jones1998efficient,
  author  = "Jones, Donald R. and Schonlau, Matthias and Welch, William J.",
  year    = 1998,
  title   = "{Efficient global optimization of expensive black-box functions}",
  journal = "Journal of Global optimization",
  volume  = 13,
  pages   = "455-492",
  publisher = "Springer"
}

@article{shahriari2015taking,
  author  = "Shahriari, Bobak and Swersky, Kevin and Wang, Ziyu and Adams, Ryan P. and De Freitas, Nando",
  year    = 2015,
  title   = "{Taking the human out of the loop: A review of Bayesian optimization}",
  journal = "Proceedings of the IEEE",
  volume  = 104,
  number  = 1,
  pages   = "148-175",
  publisher = "IEEE"
}

@article{snoek2012practical,
  author  = "Snoek, Jasper and Larochelle, Hugo and Adams, Ryan P.",
  year    = 2012,
  title   = "{Practical bayesian optimization of machine learning algorithms}",
  journal = "Advances in Neural Information Processing Systems",
  volume  = 25
}

@article{pruksawan2019prediction,
  author    = "Pruksawan, Sirawit and Lambard, Guillaume and Samitsu, Sadaki and Sodeyama, Keitaro and Naito, Masanobu",
  year      = 2019,
  title     = "{Prediction and optimization of epoxy adhesive strength from a small dataset through active learning}",
  journal   = "Science and Technology of Advanced Materials",
  volume    = 20,
  number    = 1,
  pages     = "1010-1021",
  publisher = "Taylor \& Francis"
}

@article{shields2021bayesian,
  author    = "Shields, Benjamin J. and Stevens, Jason and Li, Jun and Parasram, Marvin and Damani, Farhan and Alvarado, Jesus I. Martinez and Janey, Jacob M. and Adams, Ryan P. and Doyle, Abigail G.",
  year      = 2021,
  title     = "{Bayesian reaction optimization as a tool for chemical synthesis}",
  journal   = "Nature",
  volume    = 590,
  number    = 7844,
  pages     = "89-96",
  publisher = "Nature Publishing Group UK London"
}

@article{greenhill2020bayesian,
  author    = "Greenhill, Stewart and Rana, Santu and Gupta, Sunil and Vellanki, Pratibha and Venkatesh, Svetha",
  year      = 2020,
  title     = "{Bayesian optimization for adaptive experimental design: A review}",
  journal   = "IEEE Access",
  volume    = 8,
  pages     = "13937-13948",
  publisher = "IEEE"
}

@article{park2024efficient,
  author    = "Park, Jimin and Cheon, Mujin and Kim, David Inhyuk and Park, Daeseon and Lee, Jay H. and Koh, Dong-Yeun",
  year      = 2024,
  title     = "{Efficient extraction of hydrogen fluoride using hollow fiber membrane contactors with the aid of active-learning}",
  journal   = "AIChE Journal",
  volume    = 70,
  number    = 11,
  pages     = "e18546",
  publisher = "Wiley Online Library"
}

@article{park2024sustainable,
  author    = "Park, Jimin and Cheon, Mujin and Park, Sanghyeon and Lee, Jay H. and Koh, Dong-Yeun",
  year      = 2024,
  title     = "{Sustainable Isopropyl Alcohol Recovery via Data-Driven, Active-Learning Optimization of Vacuum Membrane Distillation}",
  journal   = "ACS Sustainable Chemistry \& Engineering",
  volume    = 12,
  number    = 31,
  pages     = "11510-11519",
  publisher = "ACS Publications"
}

@article{kandasamy2018neural,
  author  = "Kandasamy, Kirthevasan and Neiswanger, Willie and Schneider, Jeff and Poczos, Barnabas and Xing, Eric P.",
  year    = 2018,
  title   = "{Neural architecture search with bayesian optimisation and optimal transport}",
  journal = "Advances in Neural Information Processing Systems",
  volume  = 31
}

@article{ginsbourger2008discrete,
  author    = "Ginsbourger, David and Helbert, Céline and Carraro, Laurent",
  year      = 2008,
  title     = "{Discrete Mixtures of Kernels for Kriging-based optimization}",
  journal   = "Quality and Reliability Engineering International",
  volume    = 24,
  number    = 6,
  pages     = "681-691",
  publisher = "Wiley Online Library"
}

@article{roman2019experimental,
  author    = "Roman, Ibai and Santana, Roberto and Mendiburu, Alexander and Lozano, Jose A.",
  year      = 2019,
  title     = "{An experimental study in adaptive kernel selection for Bayesian optimization}",
  journal   = "IEEE Access",
  volume    = 7,
  pages     = "184294-184302",
  publisher = "IEEE"
}

@article{malkomes2018automating,
  author  = "Malkomes, Gustavo and Garnett, Roman",
  year    = 2018,
  title   = "{Automating Bayesian optimization with Bayesian optimization}",
  journal = "Advances in Neural Information Processing Systems",
  volume  = 31
}

@article{genton2001classes,
  author  = "Genton, Marc G.",
  year    = 2001,
  title   = "{Classes of kernels for machine learning: a statistics perspective}",
  journal = "Journal of Machine Learning Research",
  volume  = 2,
  number  = "Dec",
  pages   = "299-312"
}

@article{wu2019practical,
  author  = "Wu, Jian and Frazier, Peter",
  year    = 2019,
  title   = "{Practical two-step lookahead Bayesian optimization}",
  journal = "Advances in Neural Information Processing Systems",
  volume  = 32
}

@article{lam2016bayesian,
  author  = "Lam, Remi and Willcox, Karen and Wolpert, David H.",
  year    = 2016,
  title   = "{Bayesian optimization with a finite budget: An approximate dynamic programming approach}",
  journal = "Advances in Neural Information Processing Systems",
  volume  = 29
}

@article{song2022monte,
  author  = "Song, Lei and Xue, Ke and Huang, Xiaobin and Qian, Chao",
  year    = 2022,
  title   = "{Monte Carlo tree search based variable selection for high dimensional Bayesian optimization}",
  journal = "Advances in Neural Information Processing Systems",
  volume  = 35,
  pages   = "28488-28501"
}

@article{eriksson2019scalable,
  author  = "Eriksson, David and Pearce, Michael and Gardner, Jacob and Turner, Ryan D. and Poloczek, Matthias",
  year    = 2019,
  title   = "{Scalable global optimization via local Bayesian optimization}",
  journal = "Advances in Neural Information Processing Systems",
  volume  = 32
}

@article{lowe1988multivariable,
  author  = "Lowe, David and Broomhead, D.",
  year    = 1988,
  title   = "{Multivariable Functional Interpolation and Adaptive Networks}",
  journal = "Complex Systems",
  volume  = 2,
  number  = 3,
  pages   = "321--355"
}

@article{kushner1964maximum,
  author  = "Kushner, H. J.",
  year    = 1964,
  title   = "{A New Method of Locating the Maximum Point of an Arbitrary Multipeak Curve in the Presence of Noise}",
  journal = "Journal of Basic Engineering",
  volume  = 86,
  number  = 1,
  pages   = "97--106",
  doi     = "10.1115/1.3653121"
}

@article{xue2016accelerated,
  author  = "Xue, Dezhen and Balachandran, Prasanna V. and Hogden, John and Theiler, James and Xue, Deqing and Lookman, Turab",
  year    = 2016,
  title   = "{Accelerated Search for Materials with Targeted Properties by Adaptive Design}",
  journal = "Nature Communications",
  volume  = 7,
  number  = 1,
  pages   = "1--9",
  publisher = "Nature Publishing Group"
}

@article{gonzalez2024survey,
  author  = "Gonz{\'a}lez-Duque, Miguel and Michael, Richard and Bartels, Simon and Zainchkovskyy, Yevgen and Hauberg, S{\o}ren and Boomsma, Wouter",
  year    = 2024,
  title   = "{A Survey and Benchmark of High-Dimensional Bayesian Optimization of Discrete Sequences}",
  journal = "Advances in Neural Information Processing Systems",
  volume  = 37,
  pages   = "140478--140508"
}

@article{cosenza2022multi,
  author    = "Cosenza, Zachary and Astudillo, Raul and Frazier, Peter I. and Baar, Keith and Block, David E.",
  year      = 2022,
  title     = "{Multi-information Source Bayesian Optimization of Culture Media for Cellular Agriculture}",
  journal   = "Biotechnology and Bioengineering",
  volume    = 119,
  number    = 9,
  pages     = "2447--2458",
  publisher = "Wiley Online Library"
}

@inproceedings{falkner2018bohb,
  author    = "Falkner, Stefan and Klein, Aaron and Hutter, Frank",
  year      = 2018,
  title     = "{BOHB: Robust and Efficient Hyperparameter Optimization at Scale}",
  booktitle = "Proceedings of the 35th International Conference on Machine Learning (ICML-18)",
  pages     = "1437--1446",
  publisher = "PMLR"
}

@inproceedings{hoffman2011portfolio,
  author    = "Hoffman, Matthew and Brochu, Eric and De Freitas, Nando",
  year      = 2011,
  title     = "{Portfolio Allocation for Bayesian Optimization}",
  booktitle = "Proceedings of the 27th Conference on Uncertainty in Artificial Intelligence (UAI-11)",
  pages     = "327--336",
  publisher = "AUAI Press"
}

@inproceedings{vasconcelos2019no,
  author    = "Vasconcelos, Thiago de P and de Souza, Daniel A. R. M. A. and Mattos, C{\'e}sar L. C. and Gomes, Jo{\~a}o P. P.",
  year      = 2019,
  title     = "{No-PASt-BO: Normalized Portfolio Allocation Strategy for Bayesian Optimization}",
  booktitle = "Proceedings of the 31st IEEE International Conference on Tools with Artificial Intelligence (ICTAI-19)",
  pages     = "561--568",
  publisher = "IEEE"
}

@inproceedings{wilson2013gaussian,
  author    = "Wilson, Andrew and Adams, Ryan",
  year      = 2013,
  title     = "{Gaussian Process Kernels for Pattern Discovery and Extrapolation}",
  booktitle = "Proceedings of the 30th International Conference on Machine Learning (ICML-13)",
  pages     = "1067--1075",
  publisher = "PMLR"
}

@inproceedings{srinivas2010gaussian,
  author    = "Srinivas, Niranjan and Krause, Andreas and Kakade, Sham and Seeger, Matthias",
  year      = 2010,
  title     = "{Gaussian Process Optimization in the Bandit Setting: No Regret and Experimental Design}",
  booktitle = "Proceedings of the 27th International Conference on Machine Learning (ICML-10)",
  pages     = "1015--1022",
  publisher = "Omnipress"
}

@inproceedings{wilson2016deep,
  author    = "Wilson, Andrew Gordon and Hu, Zhiting and Salakhutdinov, Ruslan and Xing, Eric P",
  year      = 2016,
  title     = "{Deep Kernel Learning}",
  booktitle = "Proceedings of the 19th International Conference on Artificial Intelligence and Statistics (AISTATS 2016)",
  pages     = "370--378",
  publisher = "PMLR"
}

@inproceedings{dogan2022bayesian,
  author    = "Dogan, Vedat and Prestwich, Steven",
  year      = 2022,
  title     = "{Bayesian Optimization with Multi-Objective Acquisition Function for Bilevel Problems}",
  booktitle = "Proceedings of the 30th Irish Conference on Artificial Intelligence and Cognitive Science (AICS 2022)",
  pages     = "409--422",
  publisher = "Springer"
}

@inproceedings{tian2024boundary,
  author    = "Tian, Yunsheng and Zuniga, Ane and Zhang, Xinwei and D{\"u}rholt, Johannes P and Das, Payel and Chen, Jie and Matusik, Wojciech and Lukovic, Mina Konakovic",
  year      = 2024,
  title     = "{Boundary Exploration for Bayesian Optimization With Unknown Physical Constraints}",
  booktitle = "Proceedings of the 41st International Conference on Machine Learning (ICML 2024)",
  pages     = "48295--48320",
  publisher = "PMLR"
}

@inproceedings{eriksson2021high,
  author    = "Eriksson, David and Jankowiak, Martin",
  year      = 2021,
  title     = "{High-dimensional Bayesian optimization with sparse axis-aligned subspaces}",
  booktitle = "Proceedings of the 37th Conference on Uncertainty in Artificial Intelligence (UAI-21)",
  pages     = "493--503",
  publisher = "PMLR"
}

@inproceedings{klein2017fast,
  author    = "Klein, Aaron and Falkner, Stefan and Bartels, Simon and Hennig, Philipp and Hutter, Frank",
  year      = 2017,
  title     = "{Fast Bayesian Optimization of Machine Learning Hyperparameters on Large Datasets}",
  booktitle = "Proceedings of the 20th International Conference on Artificial Intelligence and Statistics (AISTATS 2017)",
  pages     = "528--536",
  publisher = "PMLR"
}

@inproceedings{astudillo2023qeubo,
  author    = "Astudillo, Raul and Lin, Zhiyuan Jerry and Bakshy, Eytan and Frazier, Peter",
  year      = 2023,
  title     = "{qEUBO: A Decision-Theoretic Acquisition Function for Preferential Bayesian Optimization}",
  booktitle = "Proceedings of the 26th International Conference on Artificial Intelligence and Statistics (AISTATS 2023)",
  pages     = "1093--1114",
  publisher = "PMLR"
}

@inproceedings{jimenez2023scalable,
  author    = "Jimenez, Felix and Katzfuss, Matthias",
  year      = 2023,
  title     = "{Scalable Bayesian Optimization Using Vecchia Approximations of Gaussian Processes}",
  booktitle = "Proceedings of the 26th International Conference on Artificial Intelligence and Statistics (AISTATS 2023)",
  pages     = "1492--1512",
  publisher = "PMLR"
}

@article{frazier2018tutorial,
  title   = {A Tutorial on Bayesian Optimization},
  author  = {Frazier, Peter I.},
  journal = {arXiv preprint arXiv:1807.02811},
  year    = {2018}
}

@article{chen2018bayesian,
  title   = {Bayesian Optimization in AlphaGo},
  author  = {Chen, Yutian and Huang, Aja and Wang, Ziyu and Antonoglou, Ioannis and Schrittwieser, Julian and Silver, David and de Freitas, Nando},
  journal = {arXiv preprint arXiv:1812.06855},
  year    = {2018}
}

@article{cheon2024earl,
  title   = {EARL-BO: Reinforcement Learning for Multi-Step Lookahead, High-Dimensional Bayesian Optimization},
  author  = {Cheon, Mujin and Lee, Jay H. and Koh, Dong-Yeun and Tsay, Calvin},
  journal = {arXiv preprint arXiv:2411.00171},
  year    = {2024}
}

@article{arango2021hpo,
  title   = {HPO-B: A Large-Scale Reproducible Benchmark for Black-Box HPO Based on OpenML},
  author  = {Arango, Sebastian Pineda and Jomaa, Hadi S. and Wistuba, Martin and Grabocka, Josif},
  journal = {arXiv preprint arXiv:2106.06257},
  year    = {2021}
}

@article{boyne2025bark,
  title   = {BARK: A Fully Bayesian Tree Kernel for Black-Box Optimization},
  author  = {Boyne, Toby and Folch, Jose Pablo and Lee, Robert M and Shafei, Behrang and Misener, Ruth},
  journal = {arXiv preprint arXiv:2503.05574},
  year    = {2025}
}

@article{papenmeier2025exploring,
  title   = {Exploring Exploration in Bayesian Optimization},
  author  = {Papenmeier, Leonard and Cheng, Nuojin and Becker, Stephen and Nardi, Luigi},
  journal = {arXiv preprint arXiv:2502.08208},
  year    = {2025}
}

@phdthesis{matern1960spatial,
  author  = "Mat{\'e}rn, Bertil",
  year    = 1960,
  title   = "{Spatial Variation: Stochastic Models and Their Application to Some Problems in Forest Surveys and Other Sampling Investigations}",
  type    = "{Ph.D.} diss.",
  school  = "Stockholm University",
  address = "Stockholm, Sweden"
}

@misc{simulationlib,
  title        = "Virtual Library of Simulation Experiments: Test Functions and Datasets",
  author       = "Surjanovic, S. and Bingham, D.",
  howpublished = "\url{http://www.sfu.ca/~ssurjano}",
  year         = 2025,
  note         = "Accessed: 2025-07-24"
}

@article{huang2006global,
  title={Global optimization of stochastic black-box systems via sequential kriging meta-models},
  author={Huang, Deng and Allen, Theodore T and Notz, William I and Zeng, Ning},
  journal={Journal of global optimization},
  volume={34},
  number={3},
  pages={441--466},
  year={2006},
  publisher={Springer}
}

@article{cowen2022hebo,
  title   = {Hebo: Pushing the Limits of Sample-Efficient Hyper-Parameter Optimisation},
  author  = {Cowen-Rivers, Alexander I. and Lyu, Wenlong and Tutunov, Rasul and Wang, Zhi and Grosnit, Antoine and Griffiths, Ryan Rhys and Maraval, Alexandre Max and Jianye, Hao and Wang, Jun and Peters, Jan and others},
  journal = {Journal of Artificial Intelligence Research},
  volume  = {74},
  pages   = {1269--1349},
  year    = {2022}
}

@article{balandat2020botorch,
  title   = {BoTorch: A Framework for Efficient Monte-Carlo Bayesian Optimization},
  author  = {Balandat, Maximilian and Karrer, Brian and Jiang, Daniel and Daulton, Samuel and Letham, Ben and Wilson, Andrew G. and Bakshy, Eytan},
  journal = {Advances in Neural Information Processing Systems},
  volume  = {33},
  pages   = {21524--21538},
  year    = {2020}
}

\newpage
\appendix

\section{Pseudocode for BOOST}
\label{appendix:sec:algorithm}
\begin{algorithm}[h]
\caption{Overall algorithm for BOOST}
\label{alg:boost}
\textbf{Input}: The set of kernel function candidates $\mathcal{K}$, The set of acquisition function candidates $\mathcal{A}$, Data-in-hand $D_n$ \\
\textbf{Parameter}: Constraint of initial size $(n_{\min}, n_{\max})$, ratio $r$, Maximum iterations $T_{\max}$, Stopping criterion percentile $a$ \\
\textbf{Output}: Best Kernel-Acquisition function pair ($k^*$, $\alpha^*$)

\begin{algorithmic}[1]
\STATE $n_{init} \leftarrow {clamp}(\lfloor |D_n| / r \rfloor, n_{\min}, n_{\max})$
\STATE $y_{target} \leftarrow$ top $a$ percentile of function values in $D_n$
\STATE $r_n \leftarrow$ select $n_{init}$ points by applying K-means clustering on $D_n \setminus \{(x, f(x)) \mid f(x) \le y_{target}\}$
\STATE $q_n \leftarrow D_n \setminus r_n$
\FORALL{$(k_i, \alpha_j)$ in $\mathcal{K} \times \mathcal{A}$}
    \STATE $R_0 \leftarrow r_n = \{(x_i, f(x_i))\}_{i=1}^{n_{init}}$
    \STATE $Q_0 \leftarrow \{x \mid (x, f(x)) \in q_n\}$
    \STATE $t \leftarrow 0$
    \WHILE{$t < T_{\max}$}
        \STATE Train a Gaussian Process using $k_i$ over $R_t$
        \STATE $x_{t+1} \leftarrow \arg\max_{x \in Q_t} \alpha_j(x)$
        \STATE $R_{t+1} \leftarrow R_t \cup \{(x_{t+1}, f(x_{t+1}))\}$
        \STATE $Q_{t+1} \leftarrow Q_t \setminus \{x_{t+1}\}$
        \STATE $t \leftarrow t + 1$
        \IF{$f(x_{t+1}) \le y_{target}$}
            \STATE \textbf{break}
        \ENDIF
    \ENDWHILE
    \STATE $t_{k_i,\alpha_j} \leftarrow t$
\ENDFOR
\STATE \textbf{return} $(k^*, \alpha^*) = \arg\min_{(k_i,\alpha_j)} t_{k_i,\alpha_j}$
\end{algorithmic}
\end{algorithm}

\newpage
\section{Additional Analysis of Results}
\label{appendix:sec:ranking}


\subsection{Final Performance}

\begin{figure}[htbp]
\centering
\includegraphics[width=0.85\textwidth]{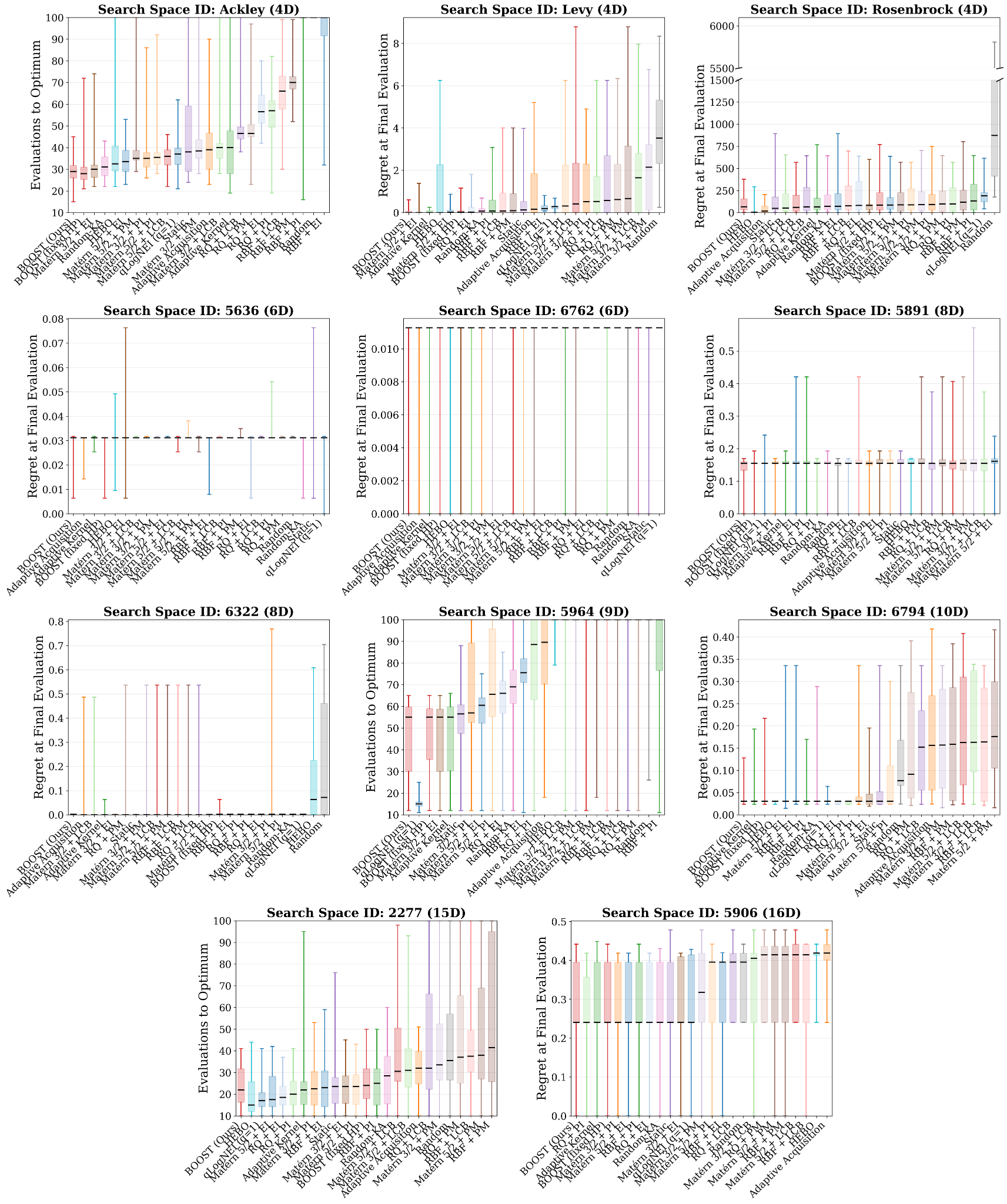}
\caption{Final performance across all benchmark tasks. For search spaces where methods reach the global optimum within 100 evaluations (Ackley, 2277, 5964), the y-axis shows the number of evaluations required to first reach the optimum (lower is better). For remaining search spaces, the y-axis shows the regret at 100 evaluations (lower is better). Box plots summarize 30 independent runs. In the Rosenbrock final-performance panel, the y-axis is broken between 1500 and 5500 for readability.}
\label{figS1:final_performance}
\end{figure}

\newpage
\subsection{Remaining Gap}

\begin{figure}[htbp]
\centering
\includegraphics[width=\textwidth]{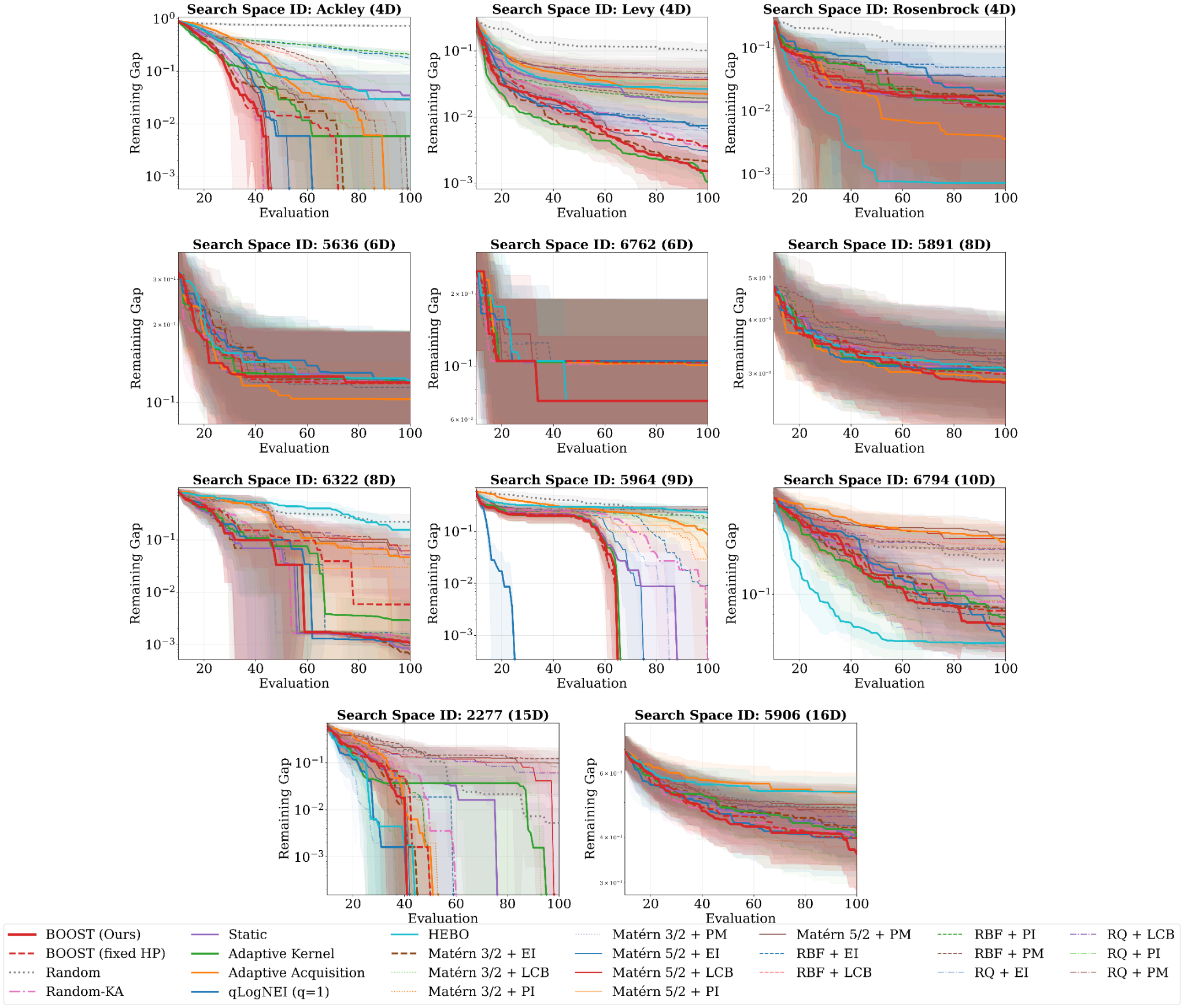}
\caption{Remaining gap across all benchmark tasks, representing the fraction of the initial regret that remains at each evaluation (lower is better). The x-axis begins at evaluation 10, after the initial design. Lines show the mean over 30 independent runs; shaded regions indicate the 95\% confidence interval.}
\label{figS2:remaining_gap}
\end{figure}

\newpage
\subsection{Worst-Case Regret}

\begin{figure}[htbp]
\centering
\includegraphics[width=\textwidth]{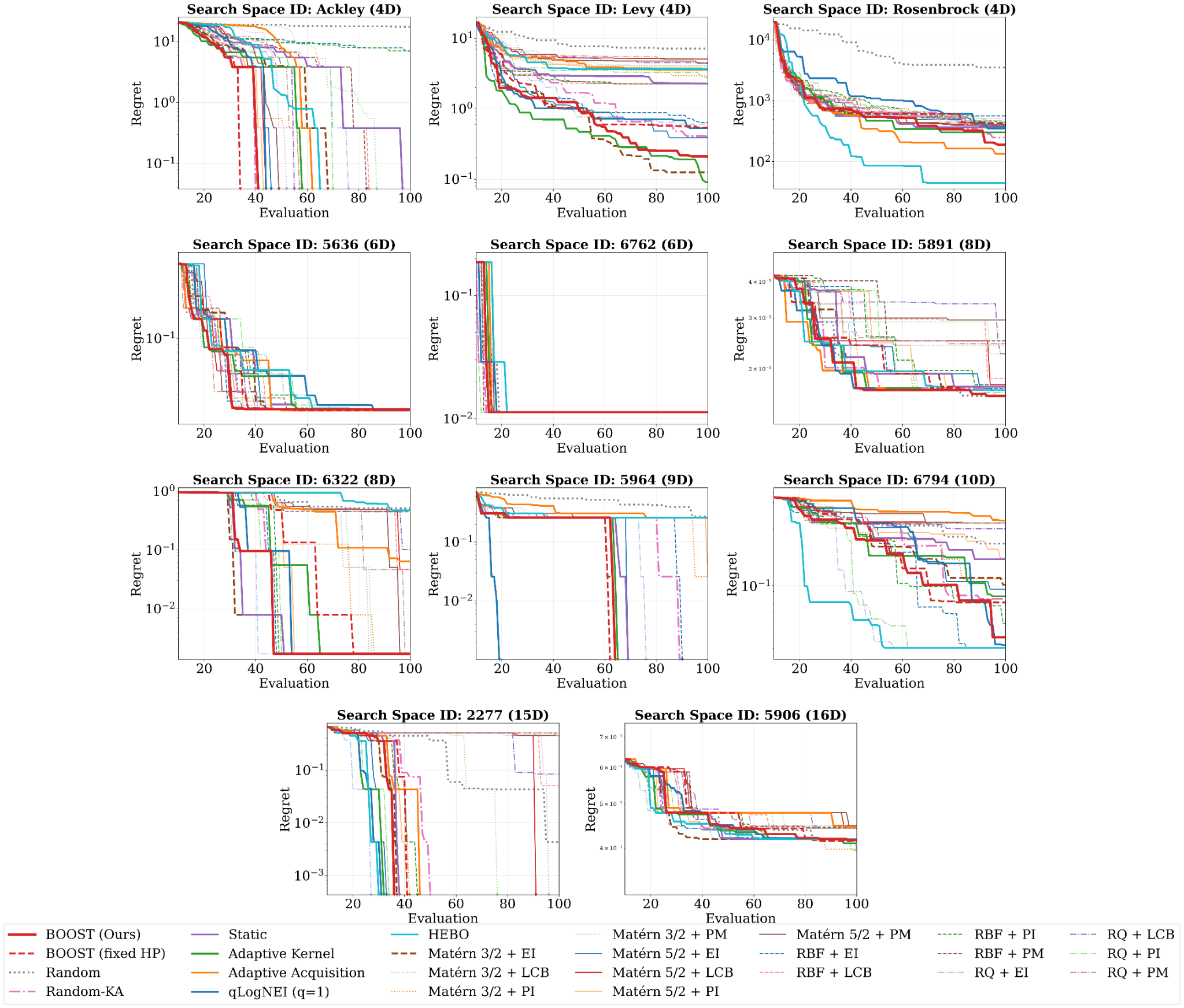}
\caption{Worst-case regret across all benchmark tasks, computed as the 90th percentile of regret over 30 independent runs (lower is better). The x-axis begins at evaluation 10, after the initial design.}
\label{figS3:worst_regret}
\end{figure}

\section{Additional Experiments and Results}
\label{appendix:sec:ablation}
In the Method section, we proposed several design strategies to enhance the robustness of BOOST:
\begin{enumerate}
    \item Using K-means clustering during the data partitioning step,
    \item Using a \( |r_n| : |q_n| \) ratio of 1:2,
    \item Using the target value \( y_{target} \) for stopping as the top 5th percentile of function values in \( D_n \),
    \item Applying a rule-based tie-breaking mechanism in the recommendation step.
\end{enumerate}
The following results validate the effectiveness of these design choices.

\subsection{Effect of Data Partitioning Method}
\label{appendix:sec:partition_method}

\begin{figure}[htbp]
\centering
\includegraphics[width=\linewidth]{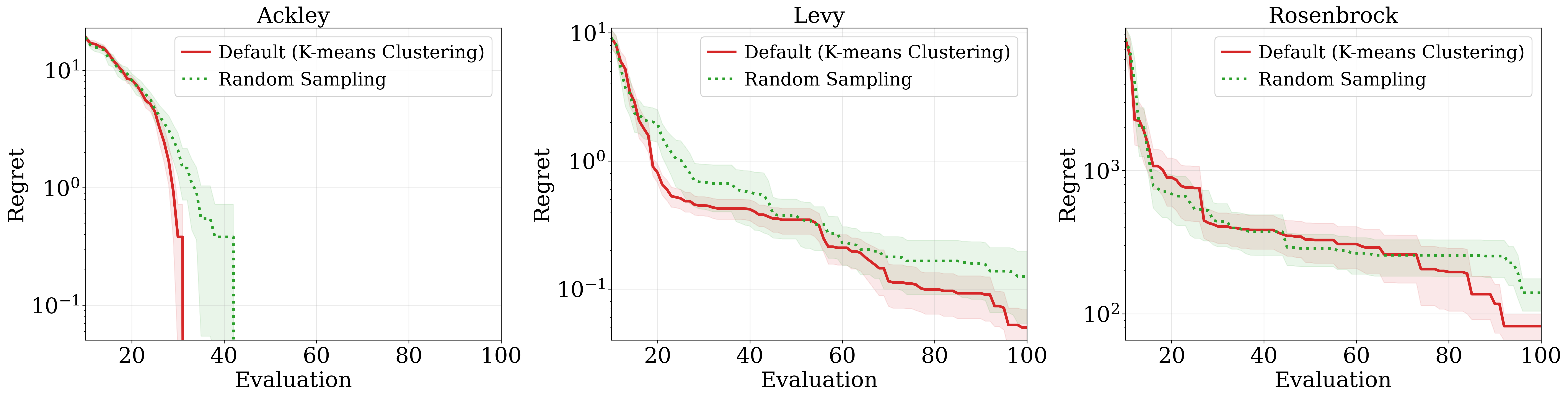}
\caption{Comparison of BOOST performance under different data partitioning strategies. The x-axis begins at evaluation 10, after the initial design}
\label{figA1:kmeans}
\end{figure}

Figure~\ref{figA1:kmeans} compares two data partitioning strategies: K-means clustering (default) and random selection of the reference subset \( r_n \) from the data-in-hand \( D_n \). BOOST with K-means clustering consistently outperformed the variant using random selection, highlighting the importance of structured data partitioning.

\subsection{Effect of Data Partitioning Size}
\label{appendix:sec:partition_size}

\begin{figure}[htbp]
\centering
\includegraphics[width=\linewidth]{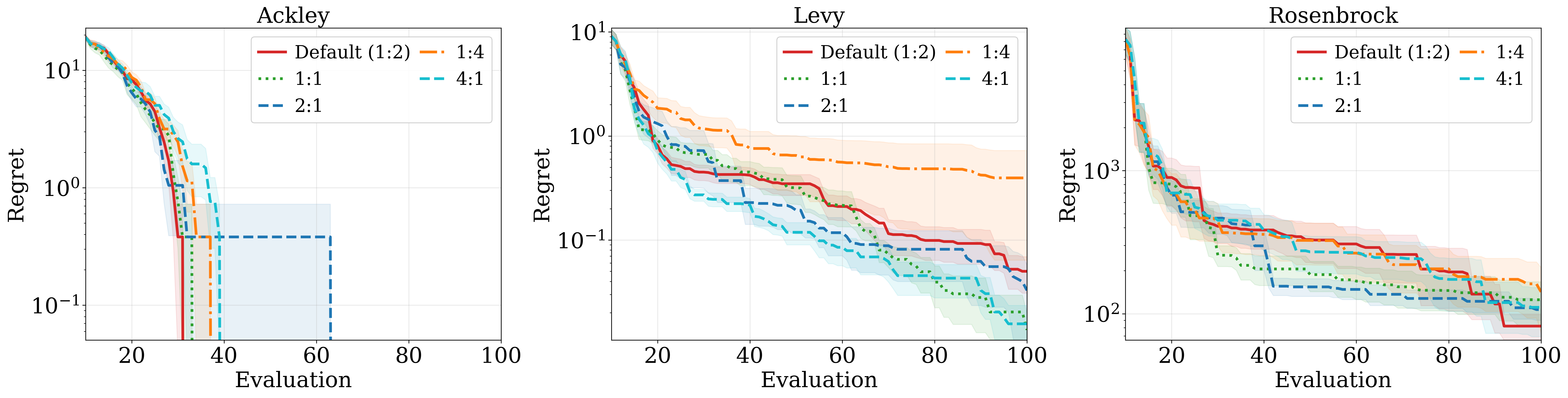}
\caption{Optimization performance of BOOST with different \( |r_n| : |q_n| \) ratios. The x-axis begins at evaluation 10, after the initial design.}
\label{figA2:ratio}
\end{figure}

Figure~\ref{figA2:ratio} compares the optimization performance of BOOST under various \( |r_n| : |q_n| \) ratios, where \( q_n \) denotes the query subset. No significant performance differences were observed across most ratios, and no particular setting consistently outperformed others. 

However, we found that using an excessively small reference set, such as a 1:4 ratio, led to a notable degradation in performance. This is likely due to the increased uncertainty of the internal Gaussian Process model, which hinders fair and stable evaluation during the optimization process.

Importantly, this robustness across different partitioning ratios demonstrates the effectiveness of BOOST's core principle: evaluating configurations based on retrospective performance within the data-in-hand provides a reliable and stable selection mechanism. The consistent performance across various ratios suggests that our approach successfully captures the inherent characteristics of different kernel--acquisition pairs, making it less sensitive to specific partitioning sizes.

\subsection{Effect of Stopping Criteria}
\label{appendix:sec:stopping}

\subsubsection{Target Percentile}
\label{appendix:sec:percentile}

\begin{figure}[htbp]
\centering
\includegraphics[width=\linewidth]{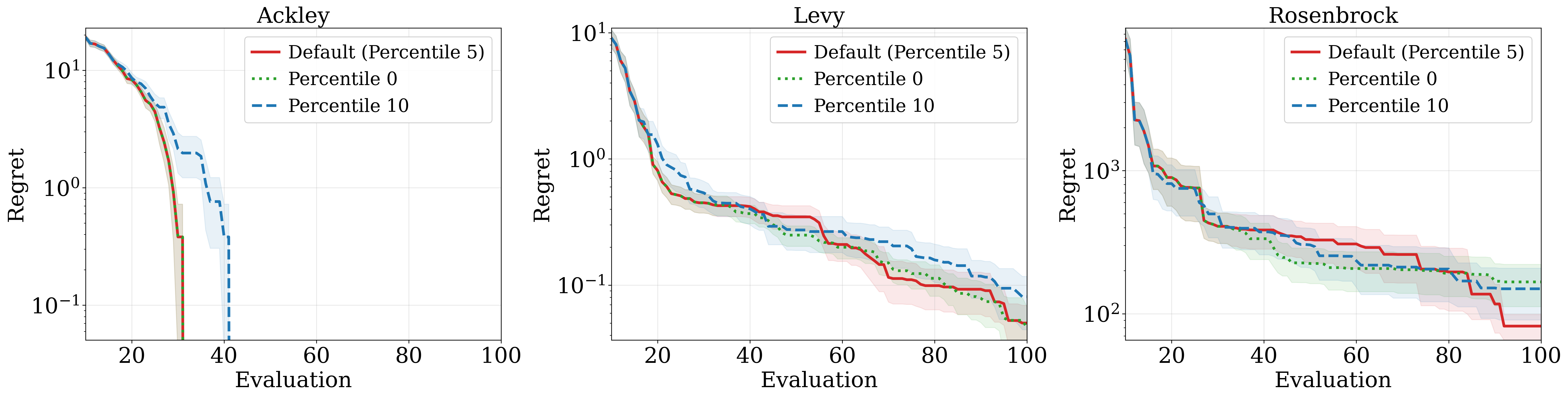}
\caption{Optimization performance of BOOST with different stopping criteria. The x-axis begins at evaluation 10, after the initial design.}
\label{figA3:percentile}
\end{figure}

Figure~\ref{figA3:percentile} compares three stopping criteria used in the
internal BO process of BOOST: setting the target value to the top 5th
percentile of $D_n$, the top 10th percentile of $D_n$, and the known global
optimum. The results demonstrate that using the top 5th percentile consistently
yields the most robust performance. The reason is that when the global optimum
is used as the target, a kernel--acquisition pair that performs well overall
may still be penalized if it fails to reach the global optimum due to
stochasticity. This can lead to an underestimation of its quality and hinder
proper selection. Conversely, setting the target too loosely, for example as
the top 10th percentile, reduces discriminatory power. As the target becomes
easier to reach, distinguishing between optimal and suboptimal strategies
becomes more difficult.

\subsubsection{Maximum Iteration}
\label{appendix:sec:max_iter}

BOOST terminates internal BO simulations either upon reaching a predefined
target value or after a fixed number of internal iterations, where the target
is defined as the top 5th percentile of observed values in the data-in-hand.
The percentile ablation in Figure~\ref{figA3:percentile} identifies the top-5\% target as the most robust stopping criterion, and the fixed horizon derived below is tied to this target rather than chosen as an independent tuning parameter.

Under this criterion, the expected number of samples required to observe a
top-5\% point under random search is 20, that is, $1/0.05$. Therefore,
kernel--acquisition strategies that do not reach the target within 20 internal
iterations are empirically indistinguishable from random search and are treated
as uninformative for retrospective comparison. If all strategies exceed this
threshold, the problem is treated as information-limited, and the strategy with
the best objective value within this range is selected.

\begin{figure}[htbp]
\centering
\includegraphics[width=\linewidth]{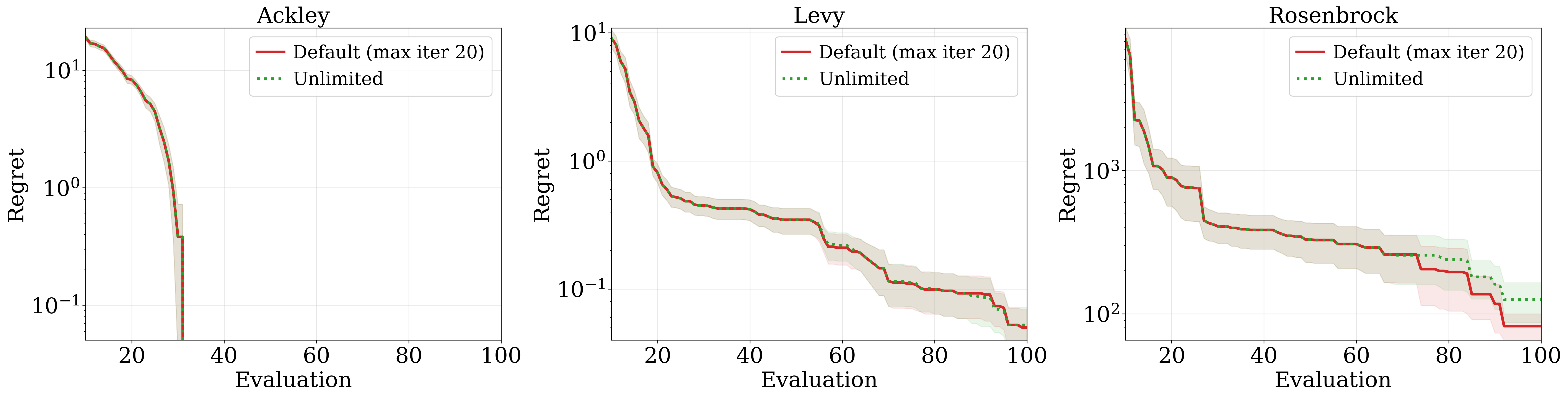}
\caption{Optimization performance of BOOST with varying maximum internal iterations. The x-axis begins at evaluation 10, after the initial design.}
\label{figA4:limit_T}
\end{figure}

We further verified that increasing the internal horizon proportionally to the
data size does not lead to qualitative changes in the behavior of BOOST, as the
internal evaluations already stabilize well within the first 20 iterations
(Figure~\ref{figA4:limit_T}).

\subsection{Effect of Tie-Breaking Rule}

\begin{figure}[htbp]
\centering
\includegraphics[width=\linewidth]{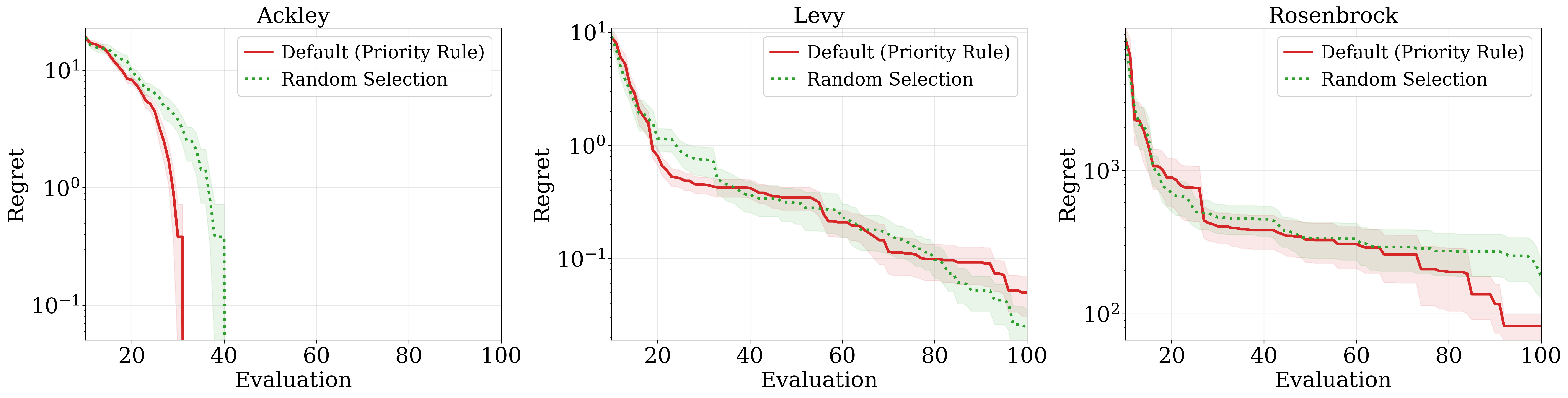}
\caption{Optimization performance of BOOST with different tie-breaking rules. The x-axis begins at evaluation 10, after the initial design.}
\label{figA5:tie-breaking}
\end{figure}

Figure~\ref{figA5:tie-breaking} compares two tie-breaking strategies used in the recommendation step: a predefined rule (see Methods section in the main text) and random selection. The results show that the predefined priority rule consistently yields better performance. This demonstrates that rule-based tie-breaking significantly impacts the outcome by acquiring more informative samples, which in turn helps to reveal the structural characteristics of the objective function.

\subsection{Effect of fixing the kernel function or acquisition function}
\label{appendix:sec:fixed_keracq}

\begin{figure}[htbp]
\centering
\includegraphics[width=\linewidth]{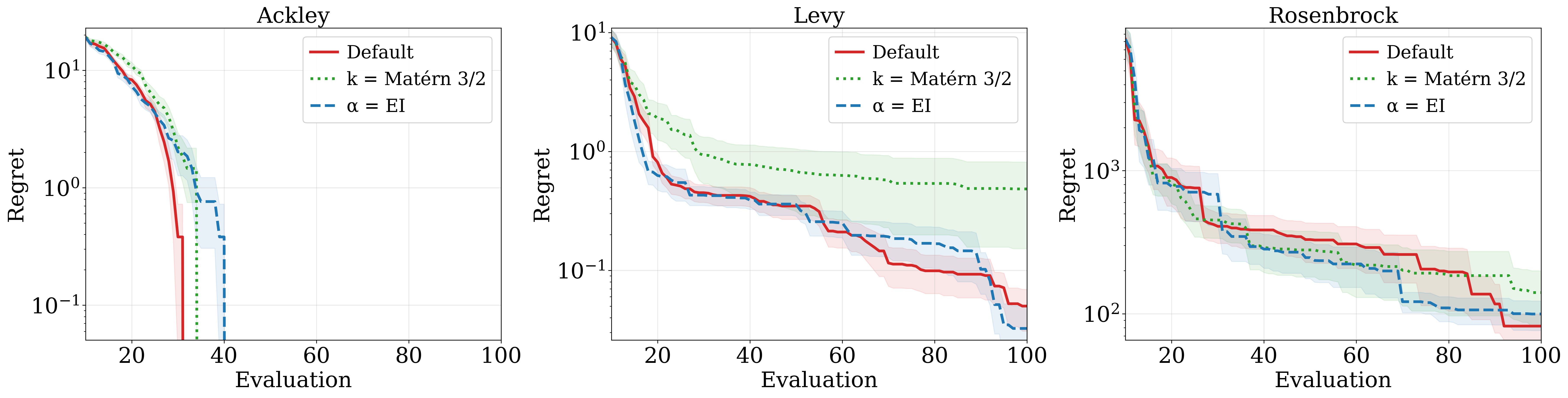}
\caption{Optimization performance of BOOST against variants in which either the kernel or the acquisition function is fixed. The x-axis begins at evaluation 10, after the initial design.}
\label{figA6:fixed_component}
\end{figure}

Figure~\ref{figA6:fixed_component} compares BOOST against variants in which
either the kernel function or the acquisition function is fixed. To ensure a
rigorous comparison, the fixed components were set to the empirically
best-performing single candidates, namely Mat\'ern 3/2 for the kernel and EI
for the acquisition function.

As shown in the results, even when one component is fixed to the globally
best-performing candidate while the other remains adaptive, the average
performance degrades compared with that of jointly adaptive BOOST. This
confirms the critical importance of jointly optimizing both the kernel and the
acquisition function.

\section{Search Space Description}

In this section, we describe the four discretized synthetic benchmark functions and the HPO-B dataset used in our experiments.

\subsection{Discretized Synthetic Benchmark Functions}
\label{appendix:sec:benchmark}

To reflect realistic discrete experimental conditions while preserving the core characteristics of each function, we discretized the domain by dividing each axis into 31--41 evenly spaced points. This maintains their optimization complexity while enabling comparison in a discrete setting. The original formulas of the benchmark functions are summarized in Table~\ref{tab:benchmarks}, which serve as the basis for our evaluation.

\begin{table}[htbp]
\caption{Synthetic Benchmark Functions}
\label{tab:benchmarks}
\begin{center}
\resizebox{\textwidth}{!}{%
\begin{tabular}{l|l}
\textbf{Function} & \textbf{Definition (with $d$: dimension)} \\
\hline
Ackley & $f(x) = -a \exp\!\left(-b \sqrt{\tfrac{1}{d} \sum x_i^2}\right) 
               - \exp\!\left(\tfrac{1}{d} \sum \cos(cx_i)\right) + a + e$, \\
       & where $a = 20,\ b = 0.2,\ c = 2\pi$. \\
Levy   & $f(x) = \sin^2(\pi w_1) + \sum_{i=1}^{d-1}(w_i - 1)^2 
               [1 + 10\sin^2(\pi w_i + 1)] + (w_d - 1)^2 
               [1 + \sin^2(2\pi w_d)]$, \\
       & where $w_i = 1 + \tfrac{x_i - 1}{4}$. \\
Rosenbrock & $f(x) = \sum_{i=1}^{d-1} \left[100(x_{i+1} - x_i^2)^2 + (1 - x_i)^2\right]$. \\
\end{tabular}
}
\end{center}
\end{table}

\begin{enumerate}
    \item \textbf{Ackley Function}

    Discretized on $x_i \in [-31.5, 31.5]$ with 37 points to preserve periodic local minima.

    \item \textbf{Levy Function}

    Discretized on $x_i \in [-10, 10]$ with 41 points to retain misleading humps leading to local minima.

    \item \textbf{Rosenbrock Function}

    Discretized on $x_i \in [-5, 10]$ with 31 points to preserve the narrow valley structure.
\end{enumerate}

\subsection{Machine Learning Hyperparameter Optimization Tasks}
\label{appendix:sec:hpob}
The HPO-B dataset \citep{arango2021hpo} is a benchmark suite providing pre-evaluated performance metrics over well-defined hyperparameter search spaces for various ML algorithms. For our experiments, we selected eight search spaces covering a range of dimensions and model types:

\begin{itemize}
    \item \textbf{Bagging + Random Forest (15D)} -- Search Space ID: 2277
    \item \textbf{Decision Trees (6D)} -- Search Space IDs: 5636, 6762
    \item \textbf{SVM (8D)} -- Search Space IDs: 5891, 6322
    \item \textbf{XGBoost (16D)} -- Search Space ID: 5906
    \item \textbf{Random Forests (9D)} -- Search Space ID: 5964
    \item \textbf{Random Forests (10D)} -- Search Space ID: 6794
\end{itemize}
Full descriptions of the search spaces are available in \citet{arango2021hpo}. All data from HPO-B are min--max normalized before evaluation. Input parameters $x$ are rounded to five decimal places (i.e., to the nearest $10^{-5}$). If multiple data points share the same $x$ after rounding, their corresponding $y$-values are averaged to obtain a single value.

\section{Information for Reproducibility}
\label{appendix:sec:reproducibility}

All experiments were repeated 30 times to ensure statistical reliability. To ensure reproducibility, we controlled randomness by adopting a consistent seeding strategy: for the $n$th trial, we used seed $n-1$ (i.e., seeds 0--29). For K-means clustering, we fixed the random seed to 42 for consistency with common practice.

\subsection{Hyperparameter Settings}

All key hyperparameters used in our experiments are listed below. Additional configuration details, such as the number of initial samples and optimization iterations, are provided in the main text. For the baseline No-PASt-BO, we used the original source code. For Best Utility, since the code was not publicly available, we implemented it ourselves based on the description in the original paper, using the same settings as BOOST to ensure a fair comparison.

\begin{itemize}
    \item[] \textbf{GP Constraints:}
    \begin{itemize}
        \item Noise constraint: \([5 \times 10^{-4}, 0.2]\)
        \item Length scale constraint: \([5 \times 10^{-6}, \sqrt{d}]\), where \(d\) is the input dimension
        \item Output scale constraint: \([0.05, 20.0]\)
    \end{itemize}

    \item[] \textbf{GP Optimizer:}
    \begin{itemize}
        \item Optimizer: Adam
        \item Learning rate: 0.05
        \item Maximum training iterations: 50
    \end{itemize}
    Following TuRBO \citep{eriksson2019scalable}, we use Adam with a fixed 50-step training budget for GP hyperparameter optimization. Our modular implementation allows additional kernels, acquisition functions, GP models, or optimizer settings to be incorporated into the candidate pool with minimal changes.
\end{itemize}

\subsection{Hardware}
All experiments were conducted under WSL2 (Windows Subsystem for Linux 2), which provides a Linux-compatible environment within the Windows operating system. The hardware used includes an AMD Ryzen 9 7900X CPU and an NVIDIA GeForce RTX 5060 Ti GPU, with 64 GB of RAM. All methods were executed on the CPU except the BoTorch default baseline, qLogNEI, which was executed on the GPU. We made this exception because qLogNEI optimizes its acquisition function over the full discrete candidate set at each BO iteration; for larger HPO-B tasks, each run can take several hours on CPU, making the full 30-seed evaluation prohibitively expensive without GPU acceleration.

\subsection{Software}
The software environment is detailed as follows:

\begin{itemize}
    \item Operating System: Ubuntu 20.04.6 LTS (via WSL2)
    \item Host OS: Windows 11
    \item Python: 3.11.11
    \item PyTorch: 2.7.1+cu128 (used on CPU for most methods and on GPU for the BoTorch default qLogNEI baseline)
    \item torchvision: 0.22.1+cu128
    \item torchaudio: 2.7.1+cu128
    \item GPyTorch: 1.14
    \item BoTorch: 0.14.0
    \item NumPy: 2.1.2
    \item SciPy: 1.15.3
    \item scikit-learn: 1.6.1
    \item pandas: 2.2.3
    \item openpyxl: 3.1.5
    \item tqdm: 4.67.1
    \item joblib: 1.5.1
\end{itemize}

For the No-PASt-BO baseline, the official code relies on the GPyOpt package, which is not compatible with recent versions. Therefore, we ran the baseline in a separate environment with older versions of NumPy/SciPy:
\begin{itemize}
    \item PyTorch: 2.7.1+cu128
    \item GPy: 1.13.2
    \item GPyOpt: 1.2.6
    \item NumPy: 1.26.4
    \item SciPy: 1.12.0
    \item pandas: 2.3.2
    \item openpyxl: 3.1.5
    \item tqdm: 4.67.1
\end{itemize}

For the BoTorch default baseline (qLogNEI), we used \texttt{qLogNoisyExpectedImprovement} with q=1q = 1
q=1, optimized over the discrete candidate set via \texttt{optimize\_acqf\_discrete}. All other settings follow BoTorch's library defaults without modification.

For the HEBO baseline, we used the official HEBO package in a separate environment because \texttt{hebo==0.3.6} requires a different numerical dependency stack from the main BoTorch/GPyTorch environment. HEBO was originally designed for continuous search spaces; to apply it under our discrete evaluation protocol, each HEBO suggestion was projected to the nearest unevaluated candidate point in the synthetic grid or HPO-B candidate set. HEBO's internal random initialization was disabled, and the same 10 seed-specific initial observations used by the other methods were supplied externally. HEBO was used in its default configuration without modifications to its surrogate or fallback behavior.
\begin{itemize}
    \item HEBO: 0.3.6
    \item PyTorch: 2.7.1+cu128
    \item NumPy: 1.24.4
    \item SciPy: 1.12.0
    \item pandas: 2.3.3
    \item openpyxl: 3.1.5
    \item tqdm: 4.67.1
\end{itemize}

\section{Scalability}

The computational complexity of BOOST is designed to remain efficient even as
the size of the data-in-hand, $D_n$, grows. Although the total dataset size may
reach the thousands, the internal simulation in BOOST uses a reference set
$r_n$, capped at 20 points, and the internal optimization horizon is likewise
limited to 20 iterations. Consequently, the Gaussian processes (GPs) used
during the internal selection phase are fit only on this small subset
(approximately 20 points), while the acquisition function is evaluated on the
remaining query set $q_n$.

To empirically assess the scalability of BOOST with respect to dimensionality,
we measured the runtime of 90-iteration optimization runs on 2D and 4D Ackley
functions (Table~\ref{tab:scalability}). We compared BOOST against a standard
baseline, Mat\'ern 3/2 + EI.

\begin{table}[htbp]
\centering
\caption{Runtime comparison over 90 optimization iterations across different dimensions.}
\label{tab:scalability}
\begin{tabular}{lccc}
\toprule
\textbf{Algorithm} & \textbf{2D Runtime (s)} & \textbf{4D Runtime (s)} & \textbf{Relative Increase (\%)} \\
\midrule
BOOST (Ours)      & 194  & 486   & +150\% \\
Mat\'ern 3/2 + EI & 21   & 117   & +460\% \\
\bottomrule
\end{tabular}
\end{table}

As shown in Table~\ref{tab:scalability}, although the absolute runtime of BOOST
is higher than that of standard EI, its relative growth in computational cost
(150\%) is substantially smaller than that of standard BO with
Mat\'ern 3/2 + EI (460\%) as dimensionality increases.
These results suggest that the computational cost of BOOST is comparatively less
sensitive to increasing dimensionality.

\section{Time Analysis}
\label{appendix:sec:time}
We compared the total runtime for 90 BO iterations on the Levy function. As described in the main text, all experiments used 10 initial samples followed by 90 optimization iterations (100 evaluations in total). The experiments were conducted on an AMD Ryzen 9 7900X CPU with 12 cores, allowing extensive parallelization across algorithms. For BOOST, although 16 candidate kernel--acquisition pairs were evaluated at each iteration, we limited the number of parallel workers to 8 (out of 12 available cores) to balance runtime efficiency and resource usage. For the other algorithms, parallel computation was applied whenever the method naturally allowed simultaneous evaluations. To preserve fidelity to the original algorithms, the code and data appendices contain results from the sequential implementations. However, in this section, we report results obtained by modifying the code to exploit parallelization for a fair comparison of runtime.

\begin{table}[htbp]
\caption{Total runtime (90 iterations) and average runtime per iteration on the Levy function.}
\label{tab:runtime}
\begin{center}
\begin{tabular}{lcc}
\textbf{Algorithm} & \textbf{Total Runtime (s)} & \textbf{Avg. per Iter. (s)} \\
\hline
BOOST (Ours)     & 608.27  & 6.76 \\
Matérn32 EI      & 163.05  & 1.81 \\
Best Utility     & 569.34  & 6.32 \\
No-PASt-BO       & 536.60  & 5.96 \\
\end{tabular}
\end{center}
\end{table}

\section{Kernel and Acquisition Function Definitions}
\label{appendix:sec:kernel_acq_defs}

The kernel functions considered in this work are defined as follows:
\begin{equation}
k_{\text{Mat\'ern}}(r) =
\frac{2^{1-\nu}}{\Gamma(\nu)}
\left( \frac{\sqrt{2\nu}\, r}{\ell} \right)^{\nu}
K_\nu\!\left( \frac{\sqrt{2\nu}\, r}{\ell} \right),
\end{equation}
\begin{equation}
k_{\mathrm{RBF}}(r) = \exp\left(-\frac{r^2}{2\ell^2}\right),
\end{equation}
\begin{equation}
k_{\mathrm{RQ}}(r) = \left(1 + \frac{r^2}{2\alpha \ell^2} \right)^{-\alpha},
\end{equation}
where $r = \|x - x'\|$ is the Euclidean distance between two input points. The Mat\'ern kernel \citep{matern1960spatial,williams2006gaussian} offers tunable smoothness ($\nu = 3/2$ and $5/2$ being common choices), the Radial Basis Function (RBF) kernel \citep{lowe1988multivariable} assumes infinite differentiability, and the Rational Quadratic (RQ) kernel \citep{williams2006gaussian} accommodates multiple characteristic length scales, with $\alpha = 2$ being widely used (we fix $\alpha = 2$ throughout).

The acquisition functions considered in this work are defined as follows:
\begin{equation}
\alpha_{\mathrm{EI}}(x) = \mathbf{E}\!\left[\max\!\big(f^- - f(x),\, 0\big)\right],
\end{equation}
\begin{equation}
\alpha_{\mathrm{PI}}(x) = \mathbf{P}\big(f(x) < f^-\big),
\end{equation}
\begin{equation}
\alpha_{\mathrm{LCB}}(x) = -(\mu(x) - \beta \sigma(x)),
\end{equation}
\begin{equation}
\alpha_{\mathrm{PM}}(x) = -\mu(x),
\end{equation}
where $f^-$ denotes the lowest observed value so far. Expected Improvement (EI) \citep{jones1998efficient} balances exploration and exploitation and is the most widely used. Probability of Improvement (PI) \citep{kushner1964maximum} considers only the probability of beating $f^-$, often leading to greedy behavior. Lower Confidence Bound (LCB) \citep{srinivas2010gaussian} controls the exploration--exploitation trade-off via the parameter $\beta$; we adopt $\beta = 0.1$ following recent studies \citep{dogan2022bayesian, tian2024boundary}, which makes it behave almost greedily \citep{papenmeier2025exploring}. Posterior Mean (PM) selects points solely based on the predicted mean, making it the most exploitative strategy that completely ignores uncertainty.

\vfill

\end{document}